\documentclass[twocolumn,letterpaper]{article}
\usepackage[review]{cvpr}      

\usepackage{graphicx}
\usepackage{amsmath}
\usepackage{amssymb}
\usepackage{booktabs}
\usepackage{multirow}
\usepackage{booktabs}
\usepackage{algorithm}
\usepackage{algpseudocode}
\usepackage{amsmath}
\usepackage[pagebackref,breaklinks,colorlinks]{hyperref}

\usepackage[capitalize]{cleveref}
\crefname{section}{Sec.}{Secs.}
\Crefname{section}{Section}{Sections}
\Crefname{table}{Table}{Tables}
\crefname{table}{Tab.}{Tabs.}

\def\cvprPaperID{*****} 
 \def\confName{CVPR}
\def\confYear{2023}

\begin{document}

\title{Progressive Open Space Expansion for Open-set Model Attribution}

\author{First Author\\
Institution1\\
Institution1 address\\
{\tt\small firstauthor@i1.org}
\and
Second Author\\
Institution2\\
First line of institution2 address\\
{\tt\small secondauthor@i2.org}
}
\maketitle

\begin{abstract}
    Model attribution aims to identify the source model of generated contents. 
    Despite its potential benefits in intellectual property protection and malicious content supervision, most existing works are under closed set scenarios, which limits the use for identifying(handling) contents generated by unseen models.
    In this work, we present the first study on \textit{Open-Set Model Attribution} (OSMA) to simultaneously attribute images to known models and identify those from unknown ones. 
    Different from existing open-set recognition (OSR) problems concentrating on semantic novelty, OSMA focuses on distinctions in visually imperceptible traces, which could hardly be handled by existing OSR methods. 
    We introduce a \underline{P}rogressive \underline{O}pen \underline{S}pace \underline{E}xpansion method (POSE) tailored for open-set model attribution, whose key idea is to simulate ``effective'' and ``diverse'' open-set samples via a progressive training mechanism. For OSMA evaluation, we construct a benchmark considering three challenging open-set setups in real-world scenarios, which includes unseen models with different training seed, architecture and dataset from seen models. Extensive experiments on the benchmark demonstrate POSE is superior to both existing fake image attribution methods and OSR methods. 
    
\end{abstract}

\section{Introduction}
\label{sec:intro}

Advanced generative modeling technology can create extremely realistic images, which also becomes a sharp weapon for disinformation production. Severe impacts of maliciously generated content have motivated a plethora of work on the detection of generated images. However, with the rapid pace of generative technology, this may not be the end. Besides detecting the generated contents, how to attribute them to their source model has been paid attention to recently. Model attribution can assist law enforcers in tracing the producers of malicious content, and it also helps claim the ownership of copyrighted generative models. However, most existing works on generated image model attribution~\cite{marra2019gans,yu2019attributing,frank2020leveraging,xuan2019scalable,yang2022aaai,bui2022repmix} are limited to a closed-set setup, which is impractical in the real world scenario where exists an ever-growing number of unseen models. In this paper, we aim at a new problem of Open-Set Model Attribution (OSMA), i.e., simultaneously attribute images to known models and identify those from unknown ones. 

\begin{figure}[h]
\begin{center}
\includegraphics[width=\linewidth]{imgs/task.png} 
\end{center}
  \caption{Open-set model recognition problem: The unknown classes include models different from known models in training seed, architecture, or training data.}   
\label{fig:intro}
\end{figure}

An intuitive way for open-set recognition (OSR) is simulating the open set space. Among existing OSR works, mixup~\cite{zhou2021learning}, GAN~\cite{neal2018open, chen2021adversarial, kong2021opengan} and copycat network~\cite{moon2022difficulty} are employed to simulate open-set samples or features. Despite their success in tackling existing OSR problems, we argue that they are sub-optimal for the aforementioned open-set model attribution problem. On the one hand, the simulated open samples are inefficient, as they may be either too close or too distant from the known distribution. For instance, counterfactual open samples~\cite{neal2018open} could hardly simulate difficult open-set space as they are already classified with low confidence, while open features generated by GANs~\cite{kong2021opengan, chen2021adversarial} tend to overlap with the known data distribution, resulting in unstable open-set discrimination without model selection by certain outliers. On the other hand, simulated open samples or features by a single generator~\cite{kong2021opengan, chen2021adversarial} or mechanism~\cite{zhou2021learning, moon2022difficulty} are not diverse enough to reduce the open space risk. This is because a single generator can hardly model the boundary of multiple known classes on a disconnected manifold. Even though a single generator could generate open images with certain semantic categories, it could only introduce a single kind of open trace left by its weights and architecture, which is inadequate for our problem. \textcolor{red}{Therefore, xxx. ADD OUR APPROACH}

In this paper, we aim to tackle the open-set model attribution problem by simulating \textit{“effective”} and \textit{“diverse”} open-set samples. As illustrated in Figure~\ref{fig:}, effectiveness guarantees the generated samples should lie on the boundary of the known distribution, and diversity ensures the generated samples vary in multiple modes, such that they could exhaustively simulate the open space.


However, it is non-trial to generate ``effective" samples. Firstly, how can we generate samples with novel low-level traces? Different from traditional OSR problems focusing on semantic novelty \textcolor{red}{like xxx}, open-set model attribution focuses on distinctions in visually imperceptible traces. 

We propose to generate open-set samples that have the same semantics as the known samples, but are different in low-level traces. To achieve this, we employ multiple tiny networks with only two convolution layers to simulate unknown traces, which reconstruct the content of known images with small residuals, modifying traces of known classes with subtle and diverse perturbations. Our empirical study proves the feasibility of utilizing tiny networks to model rich unknown traces. Secondly, how can we ensure the generated samples lie around the boundary of known classes, i.e., neither too far nor too close to the known distribution as aforementioned? We surprisingly observe that the effectiveness of generated open samples is closely related to the closed-set classification accuracy. Thus, we define the boundary of the known distribution by the mean squared error lower-bound between the generated open-set samples and the known images that have the maximum closed-set classification accuracy.

To guarantee the diversity of the generated samples, we employ multiple generators. However, naively using multiple generators may lead to a trivial solution where all generators generate images with similar traces. To avoid this, we propose a novel progressive training mechanism to span the open space iteratively. By a diversity objective, we ensure that the open samples generated by the next generator should vary largely from the samples generated by the previous generators, and be close to the known classes in the embedding space learned on all the existing samples. As such, the next generator will simulate open samples in the unexplored open space along the boundary of the known distribution. 

Our contributions can be summarized as follows:\\
1) To tackle an important challenge when applying model attribution to real-world scenarios, we are the first to raise a new issue, the open-set model attribution, which simultaneously attributes images to known models and identifies images from unknown models. \\
2) We present a novel method named POSE to generate ``effective" and ``diverse" open-set samples. We investigate in depth the boundary of known models for efficient open sample generation, and utilize a progressive training mechanism for diversity. \\
3) To validate the effectiveness of our method, we build a large dataset to represent real world open-set model attribution situations, including unseen datasets, unseen model architectures, unseen random seeds, and the combination of all. Extensive experiments on the dataset prove that POSE is superior than not only existing fake image attribution methods but also OSR methods.  


\section{Related Work}

\noindent\textbf{Open-set Recognition.} The problem of open-set recognition (OSR) aims at recognizing known classes and rejecting unknown classes simultaneously.
Existing works addressing the OSR problem mainly follow two lines: discriminative perspective and generative perspectives. 

From the discriminative perspective, many efforts have been made to adapt traditional discriminative models to the OSR problem, including works on Support Vector Machine (SVM)~\cite{scheirer2012toward, scherreik2016open, scheirer2014probability, jain2014multi}, Sparse Representation-based classifier~\cite{zhang2016sparse}, Distance-based classifier~\cite{bendale2015towards, mendes2017nearest}, Margin Distribution-based classifier~\cite{rudd2017extreme, vignotto2018extreme}, etc. Recently, Deep Neural Networks (DNN) are used in various fields owing to the excellent representation ability. Most DNN classifiers follow a typical softmax cross-entropy loss, which often leads to unreasonable high confidence on unknown samples. To solve this, Openmax~\cite{bendale2016towards} based on the Extreme Value Theory (EVT) replaces softmax for fooling/rubbish and unrelated open set images, and K-sigmoid~\cite{shu2017doc} conducted score analysis to search for threshold. 

From the generative perspective, existing works fall into two categories: 1) Employing generative models to model known-class representation~\cite{oza2019c2ae, sun2020conditional, zhang2020hybrid, guo2021conditional, perera2020generative, yoshihashi2019classification}. 2) Generating unknown samples to simulate the open space~\cite{ge2017generative, neal2018open, chen2020learning, zhou2021learning, chen2021adversarial, kong2021opengan, moon2022difficulty}. Our method falls into the second category. For this category, G-openmax~\cite{ge2017generative} improved openmax~\cite{bendale2016towards} via generating extra images to represent the unknown class. OSRCI~\cite{neal2018open} developed encoder-decoder GAN architecture to generate counterfactual examples. ARPL~\cite{chen2021adversarial} improved prototype learning~\cite{chen2020learning} by generating fake images with GAN. PROSER~\cite{zhou2021learning} simulated the open space between class boundaries based on manifold mixup~\cite{verma2019manifold}. OpenGAN~\cite{kong2021opengan} utilized GAN's discriminator as the open-set likelihood function. DIAS~\cite{moon2022difficulty} generate fakes with diverse difficulty levels by GAN and Copycat network. These proposed OSR methods mainly focuses on semantic novelty, which are sub-optimal for our problem concentrating on differences in low-level traces.


\noindent \textbf{Generated Image Attribution.} According to whether source information is injected into the generative model, fake image attribution could be classified into positive attribution~\cite{kim2020decentralized,yu2020artificial,yu2020responsible} and passive attribution~\cite{marra2019gans,yu2019attributing,frank2020leveraging,joslin2020attributing,xuan2019scalable,bui2022repmix,yang2022aaai,girish2021towards}. Works on positive attribution insert artificial fingerprint~\cite{yu2020artificial,yu2020responsible} or key ~\cite{kim2020decentralized} directly into the generative model. Then when tracing the source model, the fingerprint or key can be decoupled from generated images. Positive attribution requires ``white-box" model training and thus is limited in ``black-box" scenario when only generated images are available. What is more, positive attribution methods need to modify the model with source information, which cannot handle already trained models and may reduce the generation quality. Thus, we solve fake image attribution under the passive setting. 

Passive attribution aims at finding the intrinsic differences between different types of generated images without getting access to the generative model. The work in~\cite{marra2019gans} used averaged noise residual to represent GAN fingerprint. The work in~\cite{yu2019attributing} took the final classifier features and reconstruction residual as the image fingerprint. The work in~\cite{frank2020leveraging} observed the discrepant DCT frequency spectrums exhibited by images generated from different GAN architectures, and then used the DCT frequency spectrum for source identification. The work in~\cite{joslin2020attributing} derived a similarity metric on the frequency domain for GAN attribution. To expand the application scenario, DNA-Det~\cite{yang2022aaai} and RepMix~\cite{bui2022repmix} moved a step further from model attribution to architecture attribution. However, they still cannot deal with unseen architectures. Girish et al. ~\cite{girish2021towards} pioneer to propose an iterative pipeline to attribute images to known sources as well as discovering unknown sources. However, this work only provided a general pipeline suitable for common classification tasks and only considered unseen models trained with different architectures, neglecting the inherent challenges in open-set model attribution. Our proposed POSE is tailored for open-set model attribution, and considered more situations in open-world such as model fine-tuning, different random seeds and unseen datasets.


\section{Method}

\begin{figure}
\begin{center}
\includegraphics[width=0.99\linewidth]{imgs/method1} 
\end{center}
  \caption{Overview of progressive open-space expansion.}
\label{fig:method}
\end{figure}

\begin{figure*}
\begin{center}
\includegraphics[width=0.99\linewidth]{imgs/progressive_expansion.png} 
\end{center}
  \caption{TSNE of progressive open-space expansion.}
\label{fig:method}
\end{figure*}

\begin{figure}
\begin{center}
\includegraphics[width=0.99\linewidth]{imgs/hist-compare.png} 
\end{center}
  \caption{Hist comparison}
\label{fig:method}
\end{figure}

\begin{figure}
\begin{center}
\includegraphics[width=\linewidth]{imgs/cm-compare.png} 
\end{center}
  \caption{Similarity comparison}
\label{fig:method}
\end{figure}





We aim at the problem of open-set model attribution: to attribute an image to its source model of known classes while simultaneously recognizing images from unknown models. To solve the problem, we propose Progressive Open Space Expansion (POSE). Let $\delta\mathcal{Z}$ be the boundary of the embedding space $\mathcal{Z}$ of the known classes. $\mathcal{O}_{i}$ is the spanned open space by $i$-th augment model. The goal of POSE to simulate $\delta\mathcal{Z}$ exhaustively by the combination of $\mathcal{O}_{i}$:
\begin{equation}
\bigcup_{i}\mathcal{O}_{i} \rightarrow \delta\mathcal{Z} \label{goal}.
\end{equation}

To achieve this goal, two objectives should be satisfied:
\begin{itemize}
\setlength{\itemsep}{0pt}
\setlength{\parsep}{0pt}
\setlength{\parskip}{0pt}
\item \textbf{Efficiency objective}: the generated open space should lie on the boundary of known space, such that the generated samples are effective enough to improve the performance on challenging open-set samples.
\item \textbf{Diversity objective}: the generated open space by different generators should vary from each other, such that $\delta K$ could be simulated as large as possible.
\end{itemize}
Formally, we have the following objective functions respect to the above two objectives: 
\begin{equation}
\text{To satisfy efficiency:} \max \mathcal{O}_{i} \cap \delta\mathcal{Z} \label{eff},
\end{equation}
\begin{equation}
\text{To satisfy diversity:} \min \mathcal{O}_{i} \cap \mathcal{O}_{j}, i \ne j. \label{div}.
\end{equation}

Next, we describe the framework of POSE and how to stimulate open space by satisfying the two objectives.

\subsection{Augment Model Construction}
\label{sec:m0}

In traditional generative-based OSR methods, generative models are trained to model semantics and more sophisticated generative models result in better performance~\cite{perera2020generative}. However, the training process of generative models is resource-consuming. Different from OSR methods focusing on semantic novelty, open-set model attribution concentrates more on differences in low-level traces. Without semantics modeling requirements, we generate open samples by tiny auto-encoders with only two convolution layers. 
\noindent \textbf{Feasibility of tiny network to model unknown traces:}

\begin{figure}
\begin{center}
\includegraphics[width=1\linewidth]{imgs/cross_data_fesi2.png} 
\end{center}
  \caption{Spectral distributions of StyleGAN-Bus, StyleGAN-Cat, and generated images by sending StyleGAN-Bus images to a two convolution layer generator.}
\label{fig:feasibility}
\end{figure}

\subsection{Progressive Open Space Expansion}
\label{sec:m1}

Fig.~\ref{fig:method} shows the framework of POSE. Our model consists of task models and augment models. The task models consist of a feature extractor $F: \mathcal{X} \rightarrow \mathcal{Z}$ mapping images from input space to embedding space, and a classification head $H: \mathcal{Z} \rightarrow \mathcal{Y}$ mapping embedding to the label space. The augment models are trained progressively one by one to convert the input image $\mathbf{x}$ to augmented image $\tilde{\mathbf{x}}$ whose embedding space $\tilde{\mathcal{Z}}$ is on the boundary of $\mathcal{Z}$. 

\noindent \textbf{Training task models:} The overall loss function for task models is formulated as follows:
\begin{equation}
    \mathcal{L}_{\text{task}} = \mathcal{L}_{\text{cls}}(\mathbf{x}) + \alpha \mathcal{L}_{\text{metric}}(\mathbf{x},\tilde{\mathbf{x}}),
\end{equation}
where $\mathcal{L}_{\text{cls}}$ is the classification loss calculated on input image $\mathbf{x}$ defined in Eq.~\eqref{cls}, and $\mathcal{L}_{\text{metric}}$ is a metric loss calculated on the embedding of both input image $\mathbf{x}$ and augmented image $\tilde{\mathbf{x}}$ defined in Eq.~\eqref{metric}. $\alpha$ is a hyper-parameter.

Specifically, $\mathcal{L}_{\text{cls}}$ is a cross-entropy loss for known class classification: 
\begin{equation}
    \mathcal{L}_{\text{cls}}(\mathbf{y}, \hat{\mathbf{y}})=-\sum_i y_i \log \left(\hat{y}_i\right) \label{cls},
\end{equation}
where $\hat{\mathbf{y}}$ is the \textit{softmax} output of the classification head; $\mathbf{y}$ is the one-hot vector representing the ground truth class of $\mathbf{x}$; $y_{i}$ and $\hat{y_{i}}$ represent the $i$-th dimension of $\mathbf{y}$ and $\hat{\mathbf{y}}$, respectively. 

$\mathcal{L}_{\text{metric}}$ is a triplet loss to distinguish the embedding of augmented data from input data, and simultaneously separating different known classes and augmented classes. For an input image $x_{i}$ with class $i$, we assign the augmented image by sending it into an augmented model to a new class label $i+C$, where $C$ is the total number of known classes, getting 2$C$ classes for all input and augmented data. Let $\bar{\mathcal{X}}=\mathcal{X}\cup\tilde{\mathcal{X}}$ be the combination of input and augmented data, and $\mathcal{T}_{z}=(\bar{x}_{a}^{z},\bar{x}_{b}^{z},\bar{x}_{n}^{z})$ be one of the triplet set $\mathcal{T}$ sampled from $\bar{\mathcal{X}}$, the triplet loss is calculated as follows:
\begin{equation}
    \mathcal{L}_{\text{metric}} = \sum_{\mathcal{T}_{z} \in \mathcal{T}} [\|\bar{x}^{z}_{a} - \bar{x}^{z}_{p}\|^{2}-\|\bar{x}^{z}_{a}-\bar{x}^z_{n}\|^{2}+ m]_{+} \label{metric},
\end{equation}
where $[\cdot]_{+}=\max(0,\cdot)$ denotes the hinge loss function, and $m$ is the violate margin that requires the distance of negative pairs to be larger than the distance of positive pairs, by at least a margin $m$. 

\noindent \textbf{Training augment models:} Let $\tilde{\mathbf{x}}_{p}$ be the augment image generated by present augment model, and $\tilde{\mathbf{x}}_{o}$ is generated by previous augment models. The overall loss function for present augment model is as follows:
\begin{equation}
    \mathcal{L}_{\text{aug}} = \mathcal{L}_{\text{recons}}(\mathbf{x},\tilde{\mathbf{x}}_{p}) + \beta \mathcal{L}_{\text{eff}}(\mathbf{x}, \tilde{\mathbf{x}}_{p}) + \gamma \mathcal{L}_{\text{div}}(\tilde{\mathbf{x}}_{o},\tilde{\mathbf{x}}_{p}),
\end{equation}
where $\mathcal{L}_{\text{recons}}$ is the reconstruction loss between augmented and input image in the pixel space, $\mathcal{L}_{\text{eff}}$ is the efficiency objective for satisfying Eq.~\eqref{eff}, and $\mathcal{L}_{\text{div}}$ is the diversity objective for satisfying Eq.~\eqref{div}. $\beta$ and $\gamma$ are two hyper-parameters. The implementations of $\mathcal{L}_{\text{div}}$ and $\mathcal{L}_{\text{eff}}$ are discussed in Sec.~\ref{sec:m2}.

\subsection{Efficiency and Diversity Objectives}
\label{sec:m2}
\subsubsection{Efficiency Objective}
The efficiency objective guarantees the generated open space should lie on the boundary of known space.
In other words, the generated open space should be as close as possible to the known space, but don't overlap with it. To achieve this, we apply a hinge reconstruction loss:
\begin{equation}
    \mathcal{L}_{\text{eff}} = \max(\epsilon,\|A(\mathbf{x}) - \mathbf{x}\|),
\end{equation}
where $A$ is the present augment model. $\epsilon$ is a lower bound for the reconstruction error. In extreme case, if the weights of all convolution kernels become zero except the center element, and the center values of all input channels add up to 1, the convolution layer will completely reconstruct the input image without error, thus we apply a lower bound for reconstruction to avoid such a situation.

\noindent\textbf{Further Discussion:} For model attribution, there exist two hard cases near the known space boundary, \textit{i.e.}, models trained with only seed different, and models fine-tuned from the known models. Following the discussion in ~\cite{lukas2019deep, li2021novel, pan2022metav}, the former should be regarded as different classes as they are trained independently, and the latter should still be regarded as known classes as fine-tuning could be taken as a transformation of the original model. Luckily, open-set samples generated by our augment models naturally lies in between the two cases. 

\subsubsection{Diversity Objective}
Single augment model could only simulate one kind of trace similar to known models, which is insufficient for open space imitation. An intuitive way is to increase the number of augment models. However, as shown in Fig.~\ref{fig:tsne_wodiv}(a), naively 
increasing the number of augment models by training more models with different training seeds doesn't expand much open space, as they are quite similar in the embedding space. Thus, we expect the augmented space by different augment models should vary from each other, such that the boundary open space could be spanned as large as possible. 

We increase the diversity by a progressive training mechanism. When training the next augment model, we randomly select a previously trained augment model, and measure their distance in the embedding space of their generated images. We expect the distance could be large. However, simply increase the distance without direction would produce useless samples also far from the known classes. Hence, we apply an additional regularization between the embedding distance of augmented images and input images, to increase the diversity of augment models by optimizing toward an efficient direction.
\begin{equation}
    \mathcal{L}_{\text{div}} = - \|\tilde{\mathbf{z}}_{p} - \tilde{\mathbf{z}}_{o}\| + \|\tilde{\mathbf{z}}_{p} - \tilde{\mathbf{z}}\|
\end{equation}

\noindent\textbf{Further Discussion:} 


\subsection{Overall Training Progress}

\renewcommand{\algorithmicrequire}{\textbf{Input:}} 
\renewcommand{\algorithmicensure}{\textbf{Output:}} 
\begin{algorithm}[h]
  \caption{The proposed Progressive Open-Space Expansion (POSE)}
  \label{alg::pose}
  \begin{algorithmic}[1]
    \Require
      Image of known classes $\mathbf{x}$; 
      Number of training epochs $N$
    \Ensure
      Learned task model $M$; 
      Learned augment models $\{A_{i}\}_{i=0}^{N-1}$
    \For{$i = 0, \dots, N$} 
        \If{$i=0$}
            \State Generate augment image $\tilde{\mathbf{x}}_{p}$ using $A_{0}$
            \State Train augment model $A_{0}$ using $\mathcal{L_{\text{eff}}}(\mathbf{x},\tilde{\mathbf{x}}_{p})$
        \Else
            \State Randomly select a previously trained augment model, and generate augment image $\tilde{\mathbf{x}}_{o}$
            \State Generate augment image $\tilde{\mathbf{x}}_{p}$ using $A_{i}$
            \State Train augment model $A_{i}$ using $\mathcal{L_{\text{eff}}}(\mathbf{x},\tilde{\mathbf{x}}_{p})$ and $\mathcal{L_{\text{div}}}(\tilde{\mathbf{x}}_{o},\tilde{\mathbf{x}}_{p})$
        \EndIf
        \State Generate augment image $\tilde{\mathbf{x}}_{p}$ using $A_{i}$
        \State Train task model $M$ using $\mathcal{L_{\text{task}}}(\mathbf{x},\tilde{\mathbf{x}}_{p})$
    \EndFor
  \end{algorithmic}
\end{algorithm}

\noindent \textbf{Recognizing unknown classes:} 
Given an input test image, we feed it into the feature extractor $F$ and classification head $H$. After a $\textit{softmax}$ function, we get predicted confidence scores for each class. If the max confidence score is larger than a threshold $\theta$, the image is recognized as the known category corresponding to the index of the score. Otherwise, it is detected as unknown.

\begin{figure*}
\begin{center}
\includegraphics[width=0.8\linewidth]{imgs/progressive.png} 
\end{center}
  \caption{Progressive training mechanism}
\label{fig:progressive}
\end{figure*}

\section{Experiments}

\subsection{Experimental Details}

\begin{table*}
\small
\caption{List of dataset used in our experiments, including 4 groups: seen real, seen models, unseen real and unseen models.} 
\begin{center}
\setlength\tabcolsep{1.0pt}
\begin{tabular}{llcc}
\toprule
\multirow{3}{2cm}{\textit{\textbf{Setup I: \\Cross-seed}}} & Dataset &  Seen Models  & Unseen Models \\
\cmidrule{2-4} 
& CelebA & ProGAN\_seed0, ProGAN\_seed1 &  ProGAN\_seed2 - ProGAN\_seed9 \\
& LSUN-bedroom & ProGAN\_seed0, ProGAN\_seed1 &  ProGAN\_seed2 - ProGAN\_seed9 \\
\midrule
\multirow{3}{2cm}{\textit{\textbf{Setup II: Cross-architecture}}} & Dataset &  Seen Models  & Unseen Models\\
\cmidrule{2-4} 
& CelebA & ProGAN, MMDGAN &  StarGAN, AttGAN, SNGAN, InfoMaxGAN \\
& Faces-HQ & StyleGAN2, StyleGAN3\_r & ProGAN, StyleGAN, StyleGAN3\_t \\
& ImageNet & SAGAN, BigGAN & S3GAN, SNGAN, ContraGAN \\
& LSUN-bedroom & SNGAN, ProGAN &  MMDGAN, InfoMaxGAN \\
& Youtube & Faceshifter, FaceSwap & FSGAN, Wav2Lip \\
\midrule
\multirow{3}{2cm}{\textit{\textbf{Setup III: Cross-dataset}}} & Architectures &  Training Datasets for Seen Models  & Training Datasets for Unseen Models\\
\cmidrule{2-4} 
& StyleGAN & Bus, Cat & Airplane, Bridge, Church Outdoor, Classroom, Cow, Kitchen, Sheep \\
& StyleGAN2 & Bus, Cat & Airplane, Bridge, Church Outdoor, Classroom, Cow, Kitchen, Sheep \\
& StyleGAN3 & Bus, Cat & Airplane, Bridge, Church Outdoor, Classroom, Cow, Kitchen, Sheep \\
\bottomrule
\end{tabular}
\end{center}
\label{tab:dataset_setup}
\end{table*}

\noindent \textbf{Datasets.} We conduct extensive experiments to validate our method under various setups. Accordingly, we introduce three setups and their datasets as follows: 
\begin{itemize}
	\item[$\bullet$] \textbf{Cross-seed}: Cross-seed setup examines open-set discrimination on unseen models having the same training dataset and architecture with seen models, but trained with different training seeds. 
	Specifically, the dataset contains 20 ProGAN models trained on CelebA and LSUN-bedroom dataset, and with 10 different initalization seeds for each dataset. For the 10 models trained on CelebA or LSUN-bedroom, 2 models are randomly sampled as known classes and 8 as unknown classes.
    \item[$\bullet$] \textbf{Cross-architecture}: The goal is to evaluate on unseen models trained on the same datasets with seen models, but with different architectures. Table~\ref{tab:dataset_setup} shows a split example for the dataset, which includes 23 models trained on  
    CelebA~\cite{liu2015celeba}, Faces-HQ, ImageNet~\cite{deng2009imagenet}, LSUN-bedroom~\cite{yu2015lsun} and Faceforensics~\cite{rossler2018faceforensics}. For each dataset, 2 models with different architectures are randomly sampled as known classes and the remaining as unknown classes.
    \item[$\bullet$] \textbf{Cross-dataset}: Cross-dataset setup evaluates on unseen models trained on different datasets with seen models, but with the same architectures. Table~\ref{tab:dataset_setup} shows a split example for the dataset, which contains 44 ProGAN, StyleGAN, StyleGAN2, StyleGAN3 models. For each architecture, 11 models are trained on 11 datasets with different semantics, 2 of which are selected as known classes and the remaining as unknown classes. Known classes have the same semantics cross architectures.
\end{itemize}

\noindent \textbf{Implementation Details.} 
We describe how to train with augmented models in detail.
For experiments on open-set fake image attribution in section~\ref{sec:exp1}, 
For experiements on single model open-set recognition in section~\ref{sec:exp2}, 

\noindent \textbf{Evaluation.} We adopt accuracy to evaluate the attribution ability on the closed-set. We use the area under ROC curve (AUROC) to evaluation the known/unknown discrimination performance. We also report Open Set Classification Rate (OSCR)~\cite{dhamija2018reducing} which measures the trade-off between accuracy and open-set detection rate as a threshold on the confidence of the predicted class is varied. 


\noindent \textbf{Comparison methods.} 
We compare against four recent fake image attribution methods: Yu et al.~\cite{yu2019attributing}, DCT-CNN~\cite{frank2020leveraging}, DNA-Det~\cite{yang2022aaai}, RepMix ~\cite{bui2022repmix}, and Girish et al.~\cite{girish2021towards}. However, none of them are designed for open-set model attribution problem. For comparison, we make modifications to them as follows:
\begin{itemize}
\item[$\bullet$] Girish et al.~\cite{girish2021towards} aims at the problem of open-world GAN attribution and discovery. Rather than focusing on the open-set recognition problem to simultaneously distinguish known/unknown and recognize knowns, Girish et al.~\cite{girish2021towards} concentrates more on novel GAN discovery, which involves discovery set in the training process and violates the problem setting of OSR targeted at addressing unknown unknown classes (UUCs) (i.e., classes without any information regarding them during training). We adapt Girish et al.~\cite{girish2021towards} for open-set model attribution by extract the its WTA~\cite{} hashing based OOD detection module.
\item[$\bullet$] Yu et al.~\cite{yu2019attributing} and DCT-CNN~\cite{frank2020leveraging} are under a closed-set model attribution setup. DNA-Det~\cite{yang2022aaai} and RepMix ~\cite{bui2022repmix} are initially proposed for architecture attribution, which can also be implemented for model attribution by changing the label from architecture type to model type. We train these methods on the closed-set, and then obtain classification confidence scores on both the closed-set and open-set following the regular routine of OSR. 
\end{itemize}
We also compare with three recent state-of-the-art generative methods for OSR: OSRCI~\cite{neal2018open}, PROSER~\cite{zhou2021learning}, ARPL+CS~\cite{chen2021adversarial}, OpenGAN~\cite{kong2021opengan} and DIAS ~\cite{moon2022difficulty}. We change the backbones for all these methods to the same as our backbone for fair comparison.


\begin{table*}
\caption{Compared with existing fake image attribution methods and OSR methods.}
\begin{center}
\setlength\tabcolsep{2.0pt}
\begin{tabular}{lcccccccccccccccc}
\toprule
\multirow{2}{*}{Method} & \multicolumn{3}{c}{Cross-seed} & & \multicolumn{3}{c}{Cross-architecture} & & \multicolumn{3}{c}{Cross-dataset} & & \multicolumn{3}{c}{Cross-all} \\  
\cmidrule{2-4} \cmidrule{6-8} \cmidrule{10-12}  \cmidrule{14-16} 
 & Acc. & AUC  & OSCR  & &  Acc. & AUC  & OSCR & &  Acc. & AUC  & OSCR & &  Acc. & AUC  & OSCR \\
\midrule
PRNU et al.~\cite{} & \\
Yu et al.~\cite{yu2019attributing} &  72.73	& 57.42	& 47.22	&&  82.88	& 69.78& 	63.21	& & 84.24	& 67.57	& 61.49	& & 81.27	& 72.17	& 63.90\\
DCT-CNN~\cite{frank2020leveraging} & 85.69	& 70.73	& 65.19	&&  90.30	& 78.77	& 74.62	&&  83.51	& 71.74	& 64.81	&&  86.25	& 76.61	& 70.78\\
DNA-Det~\cite{yang2022aaai}  & \textbf{99.96}	& \textbf{91.61}	& 91.57	&& 91.68	& 78.16	& 74.08	&& \textbf{99.38}	& 48.23	& 48.08	&& 95.99	& 62.99	& 44.90\\ 
RepMix ~\cite{bui2022repmix} &  96.64 & 	67.09 &	65.96 &&	95.07 &	73.09 &	71.13 && 94.20 &	77.92 &	75.87 &&	93.14 &	76.33 & 73.72\\
\textbf{POSE} & - & 	-& -&& - & -& 	- & & 89.26&	89.25	& 81.14 & & 94.67&	89.44&	89.62\\
\bottomrule
\end{tabular}
\end{center}
\label{tab:exp1}
\end{table*}

\begin{table*}
\caption{Compared with existing fake image attribution methods and OSR methods.}
\begin{center}
\setlength\tabcolsep{2.0pt}
\begin{tabular}{lcccccccccccccccc}
\toprule
\multirow{2}{*}{Method} & \multicolumn{3}{c}{Cross-seed} & & \multicolumn{3}{c}{Cross-architecture} & & \multicolumn{3}{c}{Cross-dataset} & & \multicolumn{3}{c}{Cross-all} \\  
\cmidrule{2-4} \cmidrule{6-8} \cmidrule{10-12}  \cmidrule{14-16} 
 & Acc. & AUC  & OSCR  & &  Acc. & AUC  & OSCR & &  Acc. & AUC  & OSCR & &  Acc. & AUC  & OSCR \\
\midrule
Base  & 90.97& 	76.32& 	73.31& & 	93.20 & 	86.26& 	82.99& & 	85.04	& 68.30	& 63.88 & & 91.30 & 	79.57& 	76.85\\
Base+PROSER ~\cite{zhou2021learning}&  \\
Base+ARPL+CS ~\cite{chen2021adversarial} & 92.96 & 	78.21& 	75.57&& 	94.44& 	87.85& 	84.71&& 86.54& 	81.75& 	73.92&& 	93.12& 	80.53 &	77.86 \\
\midrule
Base+multiple AM &  95.20& 	83.42	&  82.04 & & 	96.66	& 93.42	&  90.00 & & 	87.82& 	86.38 & 77.99 & & 94.37& 	88.28& 87.63 \\
Base+multiple AM+$\mathcal{L}_{\text{eff}}$+$\mathcal{L}_{\text{div}}$  & - & 	-& -&& - & -& 	- & & 89.26&	89.25	& 81.14 & & 94.67&	89.44&	89.62\\
\bottomrule
\end{tabular}
\end{center}
\label{tab:exp1}
\end{table*}

\begin{table*}
\caption{Compared with existing fake image attribution methods and OSR methods.}
\begin{center}
\setlength\tabcolsep{2.0pt}
\begin{tabular}{lcccccccccccc}
\toprule
\multirow{2}{*}{Method} & \multicolumn{2}{c}{Cross-seed} & & \multicolumn{2}{c}{Cross-architecture} & & \multicolumn{2}{c}{Cross-dataset} & & \multicolumn{2}{c}{Cross-all} \\  
\cmidrule{2-3} \cmidrule{5-6} \cmidrule{8-9} \cmidrule{11-12}
 & Avg. Purity & NMI  && Avg. Purity & NMI &&  Avg. Purity & NMI &&  Avg. Purity & NMI \\
\midrule
Girish et al.~\cite{girish2021towards} \\
POSE w/o augmented models \\
\textbf{POSE} (full) \\
\bottomrule
\end{tabular}
\end{center}
\label{tab:exp2}
\end{table*}

\subsection{Ablation Study}
\noindent \textbf{Validation of training with augment models:}
The comparisons with and without augment models are shown in Tab.~\ref{tab:exp1}. We observe that, compared with a plain classifier, applying augment  models to simulate open space largely improves open-set model attribution performance, not only in closed-set discrimination, but also open-set recognition. The results prove that our simulated open space []. We also use t-SNE to visualize the impact of utilizing augment models in Fig.~\ref{fig:tsne}(a) and (c). We can observe that, the known space of POSE becomes closer in each class, and the distance between open and close space becomes much more distinguished.  

\noindent \textbf{Validation of $\mathcal{L}_{\text{div}}$:}
To give an intuitive understanding of how augmented models affects the distribution of generated open space, we use t-SNE to visualize with and without $\mathcal{L}_{\text{div}}$ in the embedding space. Their results are shown in Fig.~\ref{fig:tsne}(b) and (c), respectively. We observe that, with $\mathcal{L}_{\text{div}}$ added, the generated open space by different augment models vary from each other. This indicates that .
Further, we compute the similarity of open space expanded by each augment models with and without $\mathcal{L}_{\text{div}}$ as shown in the confusion matrix in Fig.~\ref{fig:cm}. The average similarity among all augmented models is [] without $\mathcal{L}_{\text{div}}$, while if $\mathcal{L}_{\text{div}}$ is not employed, the similarity decreases to [], indicating the increased diversity by introducing $\mathcal{L}_{\text{div}}$. The performances of POSE with and without $\mathcal{L}_{\text{div}}$ is presented in Tab.~\ref{tab:exp1}. We observe that with the help of $\mathcal{L}_{\text{div}}$, the open-set recognition AUC on cross-seed, cross-architecture, cross-dataset and cross-all settings improves by , , , and , respectively. The experimental results prove that the generated open space by employing $\mathcal{L}_{\text{div}}$ approaches closer to the real open space.

\noindent \textbf{Hyper-parameter tuning of $K$ and $\epsilon$:} We study the effect of two important hyper-parameters of POSE: the number of augmented models ($K$) and the reconstruction error lower-bound ($\epsilon$). We plot the open-set recognition AUC under different $K$ and $\epsilon$ in Fig.~\ref{fig:hyper} for cross-all setting. In Fig.~\ref{fig:hyper}(a), we find that the open-set recognition performance reaches the summit when $K = []$ and remain stable with $K$ increasing. This is due to the fact more augmented models could hardly expand open space near the boundary. In Fig.~\ref{fig:hyper}(b), we observe that the open-set recognition AUC reaches the summit when $\epsilon=$ and drops when $\epsilon$ continuously increases, and the closed-set classification accuracy shows a similar curve. We take the open-set samples with a $\epsilon=$ reconstruction error as ``effective" samples. This phenomenon indicates that: 1) Too close or too distant simulated open samples are either ``effective", as the former would [] and the latter is too easy to reduce open-set risk. 2) Training with effective open-set samples would also aid the learning of known samples, as the feature extractor extracts more distinguished features not only in between close/open-set but also among close-set classes.  
\subsection{Evaluation of Open-set Model Attribution}



\noindent \textbf{Comparison with fake image attribution methods:} We observe that POSE outperforms existing fake image attribution methods with a large margin in open-set recognition AUC and OSCR. RepMix~\cite{bui2022repmix} obtains better closed-set performance in cross-dataset setting but performs poorly on open set. We also compare POSE with Girish et al.~\cite{girish2021towards} to evaluate on open world GAN discovery and attribution. Results are shown in Tab.~\ref{tab:exp2}. As seen, POSE out performs Girish et al.~\cite{girish2021towards} in terms purity and NMI, which measure the clustering performance on unseen classes. These result testify the efficiency of POSE in reasoning about unseen GAN classes.

\noindent \textbf{Comparison with OSR methods:}
We compare with OSR methods by simulating open-set in Tab.~\ref{tab:exp1}, which are PROSER~\cite{zhou2021learning}, ARPL+CS~\cite{chen2021adversarial}. We report the performance of a naive cross-entropy baseline without tricks proposed in them, which is the same with POSE w/o augmented models. We observe that PROSER~\cite{zhou2021learning} and ARPL+CS~\cite{chen2021adversarial} doesn't gain much improvements compared the baseline, which proves their generated open-set samples/features are not sub-optimal for open-set model attribution. We also compare with OpenGAN~\cite{kong2021opengan} in Fig.~\ref{fig:opengan}. We observe that the performance of OpenGAN vary largely with training epochs, even decreases up to [] between two neighboring epochs, epoch [] to epoch [] for instance. The performance of OpenGAN is largely related to eval/train mode and testing batch size. These phenomenons may be due to the fact that open features generated by OpenGAN tend to overlap with known class features, such that the discriminator is unstable in distinguishing close/open set. POSE is stable to all these factors, suggesting the efficiency of its generated samples.



\section{Conclusion}

{\small
\bibliographystyle{ieee_fullname}
\bibliography{egbib}
}

\end{document}